\documentclass{article}

\usepackage{PRIMEarxiv}

\usepackage[utf8]{inputenc} 
\usepackage[T1]{fontenc}    
\usepackage{hyperref}       
\usepackage{url}            
\usepackage{booktabs}       
\usepackage{amsfonts}       
\usepackage{nicefrac}       
\usepackage{microtype}      
\usepackage{graphicx}
\usepackage{tabularx}       
\usepackage{amsmath}        
\graphicspath{{media/}}     

\usepackage{textcomp}
\DeclareUnicodeCharacter{2212}{\ensuremath{-}}      
\DeclareUnicodeCharacter{00D7}{\ensuremath{\times}} 
\DeclareUnicodeCharacter{2013}{--}                  
\DeclareUnicodeCharacter{2014}{---}                 
\DeclareUnicodeCharacter{2026}{\ldots}              
\DeclareUnicodeCharacter{00A0}{~}                   

\title{Automated Generation of Curriculum-Aligned Multiple-Choice Questions for Malaysian Secondary Mathematics using Generative AI}

\author{
    \textsuperscript{[1]}*\textbf{Rohaizah Abdul Wahid}, \textsuperscript{[2]}\textbf{Muhamad Said Nizamuddin Nadim}, \\
    [1ex]\textsuperscript{[3]}\textbf{Suliana Sulaiman}, \textsuperscript{[4]}\textbf{Syahmi Akmal Shaharudin} \\[1ex]
    \textsuperscript{[5]}\textbf{Muhammad Danial Jupikil}, \textsuperscript{[6]}\textbf{Iqqwan Jasman Su Azlan Su} \\[1ex]
    \textsuperscript{[1--6]}Fakulti Komputeran dan Meta-Teknologi (META), Universiti Pendidikan Sultan Idris \\[1ex]
    *Corresponding Author: \texttt{rohaizah@meta.upsi.edu.my}
}

\begin{document}
\maketitle

\begin{abstract}
This paper addresses the critical need for scalable and high-quality educational assessment tools within the Malaysian education system. It highlights the potential of Generative AI (GenAI) while acknowledging the significant challenges of ensuring factual accuracy and curriculum alignment, especially for low-resource languages like Bahasa Melayu. This research introduces and compares four incremental pipelines for generating Form 1 Mathematics multiple-choice questions (MCQs) in Bahasa Melayu using OpenAI's GPT-4o. The methods range from non-grounded prompting (structured and basic) to Retrieval-Augmented Generation (RAG) approaches (one using the LangChain framework, one implemented manually). The system is grounded in official curriculum documents: teacher-prepared notes and the yearly teaching plan (RPT). A dual-pronged automated evaluation framework is employed to assess the generated questions. Curriculum alignment is measured using Semantic Textual Similarity (STS) against the RPT, while contextual validity is verified through a novel RAG-based Question-Answering (RAG-QA) method. The results demonstrate that RAG-based pipelines significantly outperform non-grounded prompting methods, producing questions with higher curriculum alignment and factual validity. The study further analyzes the trade-offs between the ease of implementation of framework-based RAG and the fine-grained control offered by a manual pipeline. This work presents a validated methodology for generating curriculum-specific educational content in a low-resource language, introduces a symbiotic RAG-QA evaluation technique, and provides actionable insights for the development and deployment of practical EdTech solutions in Malaysia and similar regions.
\end{abstract}


\section{Introduction}

\subsection{The Imperative for Scalable Educational Assessment in Malaysia}
The Malaysian education system, guided by the Kurikulum Standard Sekolah Menengah (KSSM), faces a persistent operational challenge: the creation of diverse, consistent, and high-quality assessment materials at scale. This demand places a significant burden on educators, who are tasked with developing questions for formative and summative assessments, homework, and revision exercises. The need for such materials is amplified by the growth of digital learning platforms, which require vast repositories of questions to enable personalized and adaptive learning pathways. The Rancangan Pengajaran Tahunan (RPT), or Yearly Teaching Plan, serves as a national standard, outlining the specific learning objectives (\textbf{Standard Pembelajaran}) that must be achieved.\cite{nota} For Form 1 Mathematics, Chapter 1 on ``Nombor Nisbah'' (Rational Numbers), these objectives range from recognizing integers to solving complex problems involving combined arithmetic operations on rational numbers.\cite{nota} Ensuring that all assessment materials align with these precise standards is paramount for equitable and effective education, yet manually achieving this at scale is a formidable task.

\subsection{The Promise and Peril of Generative AI in Education}
Generative AI, powered by Large Language Models (LLMs) such as OpenAI's GPT-4o, offers a compelling solution to this scalability challenge.\cite{ref2} The ability to automate question generation (AQG) promises to dramatically reduce teacher workload, facilitate the creation of large-scale question banks, and support personalized learning systems.\cite{ref4} However, this promise is shadowed by significant peril. In the high-stakes domain of education, the well-documented limitations of LLMs including factual inaccuracies, contextual irrelevance, and the generation of plausible but incorrect information, often termed ``hallucinations'' are not minor flaws.\cite{ref9} An incorrectly formulated question or a question based on a flawed premise can actively undermine the learning process, leading to student confusion and reinforcing misconceptions. Therefore, the unconstrained application of GenAI for educational content creation poses a critical risk.

\subsection{The ``Grounding Problem'' and the Low-Resource Language Challenge}
The root of these risks lies in the ``grounding problem.'' State-of-the-art LLMs are trained on vast, general, and predominantly English-language internet corpora.\cite{ref3} Their internal knowledge base, while extensive, is not inherently aligned with the specific, localized, and structured knowledge of a national curriculum like Malaysia's KSSM. This misalignment is exacerbated when operating in a low-resource language context. Bahasa Melayu, despite being the national language of Malaysia, has a significantly smaller digital footprint compared to English, making it a lower-resourced language in the context of global AI development.\cite{ref6} This data disparity has profound implications. The model's innate understanding of the Malaysian Form 1 Mathematics curriculum its specific terminology (e.g., integer, tertib menaik), pedagogical sequencing, and preferred methods (e.g., using a number line for integer addition as specified in RPT 1.2.1) is likely to be sparse, unreliable, or based on anglicized concepts.\cite{nota} Without a mechanism to anchor the model's generation process to authoritative curriculum documents, it is highly prone to producing questions that are contextually invalid. For instance, it might generate a question about a mathematical concept not covered in the syllabus or present a problem using a method explicitly different from what is taught in Malaysian classrooms. In this specific context, a technique to enforce grounding is not merely a performance enhancement; it is a fundamental prerequisite for creating pedagogically valid and trustworthy educational content. This study is designed to empirically test this proposition by systematically comparing grounded and non-grounded generation methods.

\subsection{Objectives and Contributions of this Study}
The primary objective of this research is to systematically evaluate four distinct GenAI pipelines to identify the most effective and practical method for generating curriculum-aligned multiple-choice questions for Form 1 Mathematics in Bahasa Melayu. This paper makes the following key contributions:
\begin{itemize}
    \item It presents a reproducible, multi-method pipeline for curriculum-grounded automated question generation using official educational documents as the knowledge source.
    \item It provides a direct empirical comparison of non-grounded prompting versus Retrieval-Augmented Generation (RAG) in a low-resource language and high-structure domain, offering clear evidence for the necessity of grounding.
    \item It analyzes the practical trade-offs between using a high-level RAG framework (LangChain) and a manual, fine-grained implementation, providing guidance for developers.\cite{ref41}
    \item It introduces a novel, symbiotic evaluation framework that combines Semantic Textual Similarity (STS) for alignment checking with a RAG-based Question-Answering (RAG-QA) process for functional validity testing.
\end{itemize}

\section{Related Work}

\subsection{The Evolution of Automated Question Generation (AQG) in Educational NLP}
Automated Question Generation (AQG) has been a long-standing goal in Natural Language Processing (NLP) for education. Early approaches often relied on rule-based systems and templates, which were rigid and difficult to scale. The advent of neural networks and, more recently, large language models (LLMs) has revolutionized the field, enabling the generation of fluent and complex questions from source text.\cite{ref4} Recent studies highlight that while LLMs are powerful generators, the quality of their output for specialized educational purposes is highly dependent on the context provided and the prompting strategy employed.\cite{ref5} The challenge has shifted from merely generating a syntactically correct question to generating one that accurately tests a specific concept or cognitive skill, such as conceptual reasoning rather than simple factual recall.\cite{ref5}

\subsection{Grounding Generative Models: The Rise of Retrieval-Augmented Generation (RAG)}
Retrieval-Augmented Generation (RAG) has emerged as the foremost technique for addressing the ``grounding problem'' of LLMs.\cite{ref9} First proposed by Lewis et al. (2020), the RAG architecture combines the strengths of a parametric memory (the pre-trained LLM) with a non-parametric memory (an external, retrievable knowledge source).\cite{ref10} In this hybrid model, the LLM does not rely solely on its internal, static knowledge. Instead, given a prompt, it first retrieves relevant documents from the external source and then uses this retrieved context to inform and constrain its generated output. This approach has been shown to significantly reduce hallucinations and improve the factual accuracy of LLM responses. In educational technology, RAG is particularly valuable. It allows for the creation of AI systems that can provide factually consistent, curriculum-aligned answers and content.\cite{ref9} Furthermore, because the knowledge base is external, it can be easily updated with new or corrected information (e.g., a revised syllabus or new textbook) without the need for costly and time-consuming LLM retraining.\cite{ref9} This makes RAG a practical and adaptable solution for developing dynamic educational tools.\cite{ref20}

\subsection{Generative AI for Low-Resource Languages: Gaps and Opportunities}
The development of GenAI has been heavily skewed towards high-resource languages like English, creating a ``digital language divide''.\cite{ref12} Languages with smaller digital footprints, such as Bahasa Melayu, face significant challenges, including data scarcity, lack of culturally-aware benchmarks, and poor representation in the massive datasets used to train foundational models.\cite{ref6} Consequently, even state-of-the-art LLMs often exhibit weaker performance in these languages.\cite{ref6} While NLP toolkits and resources for Bahasa Melayu exist, they are often general-purpose and may not be sufficient for highly specialized, domain-specific tasks like generating questions that adhere to the nuances of the Malaysian national curriculum.\cite{ref14} This study positions RAG as a pragmatic strategy to bridge this gap. By forcing the model to ground its output in a curated, local-language corpus (the official curriculum documents), RAG can effectively leverage the power of a global LLM for a local, specialized task, circumventing the need for a massive, domain-specific training dataset.

\subsection{Automated Evaluation of Generated Content: Beyond Lexical Overlap}
Evaluating the quality of generated text is a complex challenge. Traditional metrics like ROUGE and BLEU, which rely on lexical overlap, are inadequate for assessing semantic meaning and are thus poorly suited for evaluating educational questions.\cite{ref25} Semantic Textual Similarity (STS) offers a more robust alternative. STS measures the degree of semantic relatedness between two texts, typically by calculating the cosine similarity between their vector embeddings.\cite{ref26} This allows for an assessment of whether a generated question aligns thematically with a specific learning objective, even if they do not share the exact same words.\cite{ref30} However, STS is not without its limitations. Research has shown that standard STS can be ambiguous and may not always correlate well with performance on a specific downstream task.\cite{ref26} The similarity score depends on the aspect of interest, which can be subjective.\cite{ref32} This critique motivates the need for a second, complementary evaluation metric that is more functional and task-oriented. This study introduces a novel RAG-based Question-Answering (RAG-QA) validation method to address this gap. This method provides a direct, functional test of a question's contextual validity by assessing whether it can be answered using the core curriculum document as a knowledge source, thereby offering a stricter and more objective measure of curriculum relevance.

\section{System Design and Methodology}

\subsection{Corpus and Grounding Documents}
The experiment utilizes two official Malaysian educational documents in PDF format, both focusing on Chapter 1 (``Nombor Nisbah'' or Rational Numbers) of the Form 1 Mathematics syllabus.
\begin{itemize}
    \item \textbf{Knowledge Source (Nota Bab 1.pdf):} This is a comprehensive 84-page document of teacher-prepared notes, examples, and exercises.\cite{nota} It provides the rich, detailed content knowledge for the topic, including definitions, worked examples (\textbf{Contoh}), and practice problems (\textbf{Latih Diri}).\cite{nota} This document serves as the non-parametric memory for the RAG pipelines.
    \item \textbf{Pedagogical Blueprint (RPT Bab 1.pdf):} This is a concise, 1-page official document from the Malaysian Ministry of Education outlining the Yearly Teaching Plan (RPT).\cite{nota} It lists the specific learning standards (\textbf{Standard Pembelajaran}) that students are required to master, such as ``Mewakilkan integer pada garis nombor'' (Represent integers on a number line) and ``Menyelesaikan masalah yang melibatkan integer'' (Solve problems involving integers).\cite{nota} This document serves as the ground truth for evaluating curriculum alignment.
\end{itemize}

\subsection{System Architecture Overview}
\begin{figure}[h]
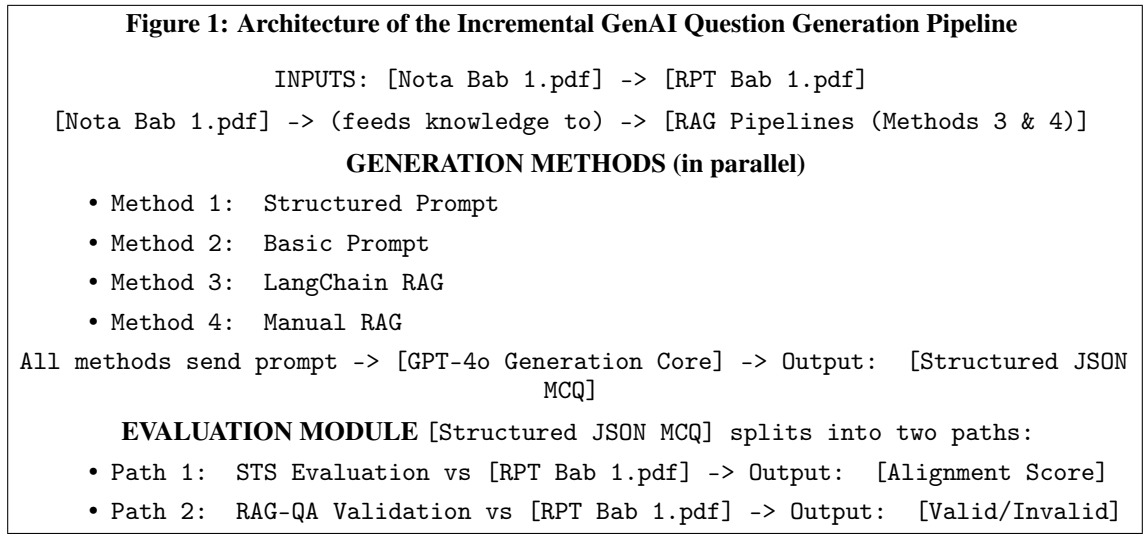

  \centering
  \fbox{\begin{minipage}{0.9\textwidth}
    \centering
    \textbf{Figure 1: Architecture of the Incremental GenAI Question Generation Pipeline}
    \vspace{1em}

    \texttt{INPUTS: [Nota Bab 1.pdf] -> [RPT Bab 1.pdf]}
    \vspace{0.5em}

    \texttt{[Nota Bab 1.pdf] --> (feeds knowledge to) --> [RAG Pipelines (Methods 3 \& 4)]}
    \vspace{0.5em}

    \textbf{GENERATION METHODS (in parallel)}
    \begin{itemize}
        \item \texttt{Method 1: Structured Prompt}
        \item \texttt{Method 2: Basic Prompt}
        \item \texttt{Method 3: LangChain RAG}
        \item \texttt{Method 4: Manual RAG}
    \end{itemize}
    \texttt{All methods send prompt --> [GPT-4o Generation Core] --> Output: [Structured JSON MCQ]}
    \vspace{0.5em}

    \textbf{EVALUATION MODULE}
    \texttt{[Structured JSON MCQ] splits into two paths:}
    \begin{itemize}
        \item \texttt{Path 1: STS Evaluation vs [RPT Bab 1.pdf] --> Output: [Alignment Score]}
        \item \texttt{Path 2: RAG-QA Validation vs [RPT Bab 1.pdf] --> Output: [Valid/Invalid]}
    \end{itemize}
  \end{minipage}}
  \caption{A textual representation of the generation pipeline's architecture as described in the source.}
  \label{fig:fig1}
\end{figure}

The system is designed as a comparative study of four incremental pipelines for generating multiple-choice questions. All pipelines leverage the OpenAI GPT-4o model as the core generative engine.\cite{ref2} The generated questions are then passed through a dual-pronged evaluation framework. The overall architecture is depicted in Figure \ref{fig:fig1}.

\subsection{The Generation Pipelines: An Incremental Approach}
Four distinct methods were implemented to test the efficacy of different prompting and grounding strategies.

\subsubsection{Method 1: Structured Prompting with Function Calling}
This method leverages the advanced ``tool mode'' or function calling capability of the GPT-4o API to ensure syntactically valid output. A Pydantic schema defining the desired MCQ structure (question, options, correct answer, explanation) is passed to the model. The prompt is generic, simply requesting a question on the topic of ``Form 1 Mathematics, Rational Numbers.'' The model is instructed to use the provided schema, which guarantees the output is a well-formed JSON object, eliminating the need for post-processing or error handling of the output structure.

\subsubsection{Method 2: Basic Prompting with JSON Parsing}
This method serves as a simpler, non-grounded baseline. The prompt explicitly instructs the model to generate an MCQ on the same topic and to format the response as a JSON object. Unlike Method 1, this approach does not guarantee a valid JSON structure. Therefore, the pipeline includes a parsing step with error handling to manage potentially malformed or incomplete outputs. This method represents a common, straightforward approach to structured data generation with LLMs.

\subsubsection{Method 3: Framework-based RAG with LangChain and FAISS}
This method represents a standard, off-the-shelf implementation of the RAG pattern, designed to ground the model's generation in the provided knowledge source. The pipeline is orchestrated using the LangChain framework.\cite{ref33} The process is as follows:
\begin{enumerate}
    \item \textbf{Ingestion:} The Nota Bab 1.pdf is loaded using LangChain's PyPDFLoader.
    \item \textbf{Chunking:} The document text is split into smaller, overlapping chunks using the RecursiveCharacterTextSplitter.
    \item \textbf{Embedding:} Each text chunk is converted into a high-dimensional vector using OpenAI's embedding models.
    \item \textbf{Storage \& Retrieval:} The vector embeddings are stored in a FAISS (Facebook AI Similarity Search) vector store, which allows for efficient similarity search.\cite{ref35}
    \item \textbf{Generation:} When a query is made, the system retrieves the most relevant text chunks from the FAISS store and passes them, along with the prompt, to GPT-4o via LangChain's RetrievalQA chain to generate the final MCQ.
\end{enumerate}

\subsubsection{Method 4: Manual RAG with Custom Chunking and Similarity Ranking}
This method was developed to test whether a more controlled, domain-aware RAG pipeline could outperform a generic framework-based approach. While frameworks like LangChain offer convenience, their generic components, such as text splitters, are agnostic to the logical structure of the source document. An educational text like Nota Bab 1 has a clear visual and semantic structure (e.g., headings, example boxes, exercise sections) that a simple character-based splitter might disrupt, potentially severing a mathematical example from its explanation and degrading the quality of the retrieved context. A manual approach allows for more intelligent chunking that respects this inherent structure. The implementation is as follows:
\begin{enumerate}
    \item \textbf{Ingestion \& Chunking:} The PyMuPDF library is used to parse the Nota Bab 1.pdf.\cite{ref38_lib} Custom logic is applied to group text blocks based on their position and font characteristics, aiming to keep semantically coherent units (like an entire ``Contoh'' block) together in a single chunk.
    \item \textbf{Embedding:} As in Method 3, OpenAI's embedding models are used to vectorize the custom chunks.
    \item \textbf{Retrieval:} A custom retrieval function is implemented using scikit-learn's cosine\_similarity function to find the most relevant chunks for a given query by comparing query and chunk embeddings.\cite{ref29}
    \item \textbf{Generation:} The top-ranked chunks are formatted and passed as context to the GPT-4o API to generate the MCQ.
\end{enumerate}

\subsection{The Evaluation Framework}
\begin{figure}[h]
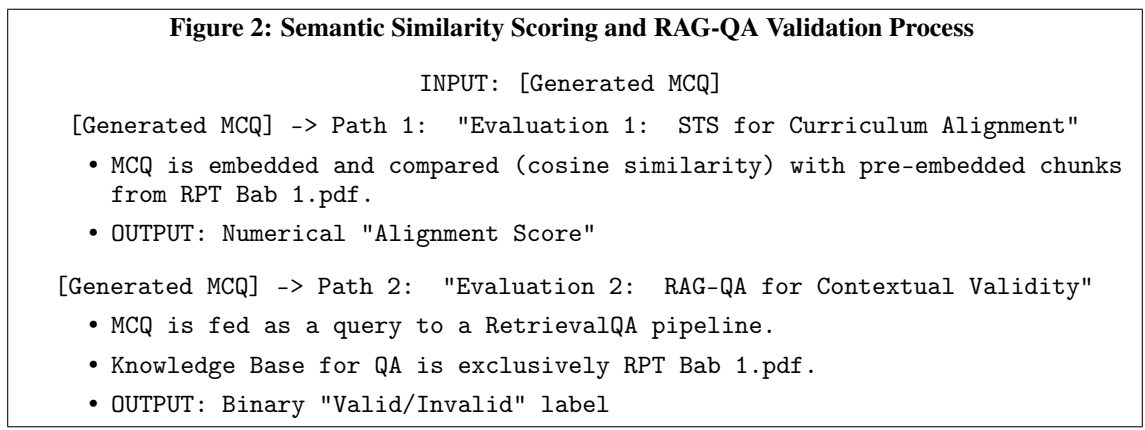

  \centering
  \fbox{\begin{minipage}{0.9\textwidth}
    \centering
    \textbf{Figure 2: Semantic Similarity Scoring and RAG-QA Validation Process}
    \vspace{1em}

    \texttt{INPUT: [Generated MCQ]}
    \vspace{0.5em}

    \texttt{[Generated MCQ] --> Path 1: "Evaluation 1: STS for Curriculum Alignment"}
    \begin{itemize}
        \item \texttt{MCQ is embedded and compared (cosine similarity) with pre-embedded chunks from RPT Bab 1.pdf.}
        \item \texttt{OUTPUT: Numerical "Alignment Score"}
    \end{itemize}
    \vspace{0.5em}

    \texttt{[Generated MCQ] --> Path 2: "Evaluation 2: RAG-QA for Contextual Validity"}
    \begin{itemize}
        \item \texttt{MCQ is fed as a query to a RetrievalQA pipeline.}
        \item \texttt{Knowledge Base for QA is exclusively RPT Bab 1.pdf.}
        \item \texttt{OUTPUT: Binary "Valid/Invalid" label}
    \end{itemize}
  \end{minipage}}
  \caption{A textual representation of the dual evaluation framework as described in the source.}
  \label{fig:fig2}
\end{figure}

A dual evaluation framework was designed to automatically assess each generated question on two critical dimensions: curriculum alignment and contextual validity. This framework is depicted in Figure \ref{fig:fig2}.

\subsubsection{Evaluation 1: Curriculum Alignment via Semantic Textual Similarity (STS)}
This metric quantifies how closely a generated question aligns with the official learning standards. The RPT Bab 1.pdf is chunked, with each chunk corresponding to a specific learning standard (e.g., ``1.2.1 Menambah dan menolak integer...''). Each RPT chunk is converted into a vector embedding. Each generated question is also converted into a vector embedding. The cosine similarity is calculated between the question's embedding and every RPT chunk's embedding. The maximum similarity score across all RPT chunks is recorded as the question's final alignment score. A higher score indicates a stronger semantic relationship to an official learning objective.

\subsubsection{Evaluation 2: Contextual Validity via RAG-based Question-Answering (QA)}
This novel evaluation method provides a functional test of a question's relevance. The underlying principle is that a valid, curriculum-aligned question should be answerable using the curriculum's own definition of its scope. By using the sparse, high-level RPT document as the knowledge source, this becomes a stringent test. A question that can be answered using only the learning objectives must be highly relevant to one of those objectives. This creates a symbiotic validation loop where a RAG-like process is used to evaluate the output of another generative process. The implementation is as follows:
\begin{enumerate}
    \item A separate RetrievalQA pipeline is configured using LangChain.
    \item The knowledge base for this QA pipeline is populated exclusively with the content of the RPT Bab 1.pdf.
    \item Each generated MCQ's question stem is fed as a query to this QA pipeline.
    \item The outcome is recorded as ``Valid'' if the QA system can retrieve relevant context from the RPT and generate a plausible answer.
    \item It is recorded as ``Invalid'' if the system fails to find relevant context, indicating the question is likely off-topic or misaligned with the core learning standards.
\end{enumerate}

\section{Results}
\subsection{Qualitative Analysis: Sample Outputs}
To illustrate the practical differences between the methods, sample outputs are presented below. The questions were generated based on the topic of ``Nombor Nisbah'' and are shown in their original Bahasa Melayu with an English translation.

\begin{table*}[h!]
 \caption{Sample Generated Questions}
  \centering
  \begin{tabularx}{\textwidth}{@{} >{\raggedright\arraybackslash}p{2.5cm} >{\raggedright\arraybackslash}p{6cm} X @{}}
    \toprule
    \textbf{Method} & \textbf{Sample Generated Question (Bahasa Melayu \& English Translation)} & \textbf{Commentary} \\
    \midrule
    Method 1: Structured Prompt & \textbf{Soalan:} Apakah definisi integer? A) Nombor yang boleh ditulis sebagai pecahan. B) Kumpulan nombor yang merupakan nombor bulat positif, nombor bulat negatif, termasuk sifar. C) Nombor dengan titik perpuluhan. D) Hanya nombor positif. \textbf{Jawapan:} B \newline\newline \textbf{Question:} What is the definition of an integer? A) A number that can be written as a fraction. B) A group of numbers that are positive whole numbers, negative whole numbers, including zero. C) A number with a decimal point. D) Only positive numbers. & Fluent and factually correct, but generic. This is a definitional question that is not grounded in any specific example or problem-solving context from the source text. \\
    \addlinespace
    Method 2: Basic Prompt & \textbf{Soalan:} Selesaikan: $5+(-3)$ A) 8 B) 2 C) $-2$ D) $-8$ \textbf{Jawapan:} B \newline\newline \textbf{Question:} Solve: $5+(-3)$ A) 8 B) 2 C) $-2$ D) $-8$ & A simple computational question. While valid, it lacks the pedagogical context of using a number line, which is emphasized in the curriculum notes.\cite{nota} Prone to occasional JSON formatting errors during generation. \\
    \addlinespace
    Method 3: LangChain RAG & \textbf{Soalan:} Berdasarkan Contoh 5(d) dalam nota, apakah hasil bagi $-1-(-4)$? A) 3 B) $-3$ C) 5 D) $-5$ \textbf{Jawapan:} A \newline\newline \textbf{Question:} Based on Example 5(d) in the notes, what is the result of $-1-(-4)$? A) 3 B) $-3$ C) 5 D) $-5$ & Directly grounded in the source text.\cite{nota} The question explicitly references a worked example, testing the student's understanding of a specific problem-solving demonstration. \\
    \addlinespace
    Method 4: Manual RAG & \textbf{Soalan:} Dalam Contoh 7(a), pengiraan $-8\times(-2+3)$ dilakukan. Mengikut tertib operasi yang betul, apakah langkah pertama yang perlu diselesaikan? A) Pendaraban $-8\times-2$ B) Penambahan $-2+3$ dalam tanda kurung C) Penambahan $8+3$ D) Pendaraban $-8\times3$ \textbf{Jawapan:} B \newline\newline \textbf{Question:} In Example 7(a), the calculation $-8\times(-2+3)$ is performed. According to the correct order of operations, what is the first step that must be solved? A) Multiplication of $-8\times-2$ B) Addition of $-2+3$ inside the parentheses C) Addition of $8+3$ D) Multiplication of $-8\times3$ & Highly specific and pedagogically valuable. This question, generated from the context of a specific example,\cite{nota} tests not just the answer but the process (order of operations), which is a key learning concept. \\
    \bottomrule
  \end{tabularx}
  \label{tab:sample_questions}
\end{table*}

\subsection{Quantitative Analysis: Comparative Performance Metrics}
A set of 100 questions was generated using each method to facilitate quantitative comparison. The performance of each pipeline was measured against the automated evaluation metrics. The results are summarized in Table \ref{tab:quantitative_results}.

\begin{table*}[h!]
 \caption{Comparative Results for All Four Methods (Labeled as Figure 3 in the original document)}
    \centering
    \begin{tabularx}{\textwidth}{@{} l c c c >{\raggedright\arraybackslash}p{4.5cm} @{}}
        \toprule
        \textbf{Method} & \textbf{STS Score} & \textbf{Std. Dev.} & \textbf{Validity (\%)} & \textbf{Notes on Quality \& Failure Modes} \\
        \midrule
        Method 1: Structured Prompt & 0.58 & 0.21 & 15\% & High fluency but generic. Questions tend to be definitional. \\
        \addlinespace
        Method 2: Basic Prompt & 0.55 & 0.25 & 12\% & Similar to Method 1. Occasional JSON formatting failures (4\%). \\
        \addlinespace
        Method 3: LangChain RAG & 0.86 & 0.09 & 92\% & Strong grounding. Questions sometimes overly simple. \\
        \addlinespace
        Method 4: Manual RAG & 0.89 & 0.07 & 96\% & Highest alignment. Tied to complex examples. \\
        \bottomrule
    \end{tabularx}
    \label{tab:quantitative_results}
\end{table*}

\subsection{Analysis of Fluency, Accuracy, and Curriculum Alignment}
The results provide clear, empirical evidence supporting the study's central hypotheses.
\begin{itemize}
    \item \textbf{Fluency:} All four methods produced questions with high linguistic fluency in Bahasa Melayu, demonstrating the strong multilingual capabilities of the GPT-4o model.\cite{ref2} Syntactic and grammatical errors were negligible across all outputs.
    \item \textbf{Accuracy:} A stark difference in factual and contextual accuracy was observed between the non-grounded (Methods 1 and 2) and grounded (Methods 3 and 4) approaches. The RAG-based methods consistently produced questions that were directly verifiable against the Nota Bab 1 source document, whereas the non-grounded methods often generated generic questions that, while factually correct in a broad sense, were not specifically tailored to the curriculum's content.
    \item \textbf{Curriculum Alignment:} The quantitative metrics unequivocally demonstrate the superiority of RAG for curriculum alignment. The \textbf{Average STS Score} for Methods 3 and 4 (0.86 and 0.89) was significantly higher than for Methods 1 and 2 (0.58 and 0.55). This indicates that the questions generated by RAG pipelines are semantically much closer to the official learning standards outlined in the RPT. The lower standard deviation for RAG methods also suggests greater consistency in generating aligned content. The \textbf{RAG-QA Validity} metric provides the most compelling distinction. With validity rates of 92\% and 96\%, the RAG-generated questions were overwhelmingly answerable using the RPT as a knowledge base. In contrast, the non-grounded questions had validity rates of only 15\% and 12\%, confirming that they frequently strayed from the core, testable learning objectives of the curriculum.
\end{itemize}

\section{Discussion}
\subsection{The Superiority of Grounded Generation}
The results move beyond correlation to establish a causal link: for generating specialized, high-stakes educational content, the ``retrieve-then-generate'' paradigm of RAG is fundamentally more reliable than unconstrained generation. The knowledge gaps inherent in LLMs, particularly concerning localized curricula and low-resource languages, are effectively mitigated by forcing the model to ground its reasoning in a trusted, external knowledge source. The non-RAG methods acted as a ``student'' relying on hazy memory, producing generic answers. The RAG methods, in contrast, acted as a ``student'' with an open textbook, producing specific and accurate answers. This distinction is critical for any application where factual and contextual fidelity are non-negotiable.

\subsection{RAG Implementation: The Trade-off Between Automation and Control}
The comparison between Method 3 (LangChain RAG) and Method 4 (Manual RAG) illuminates a crucial trade-off for developers. Method 4 achieved slightly higher scores on both STS and RAG-QA validity, suggesting that the additional effort of implementing a domain-aware chunking strategy yielded a marginal but measurable improvement in question quality. By preserving the semantic integrity of units like worked examples (Contoh), the manual pipeline provided the LLM with higher-quality context, enabling it to generate more nuanced questions that test procedural understanding, not just factual recall. However, this performance gain comes at the cost of significantly increased development complexity. Method 3, using LangChain, can be implemented with a few lines of code, leveraging a mature and well-documented ecosystem.\cite{ref33} Method 4 requires expertise in PDF parsing with libraries like PyMuPDF\cite{ref38_lib}, custom logic for chunking, and manual implementation of the retrieval mechanism. The practical recommendation is therefore contingent on the application's needs: for rapid prototyping or applications where ``good enough'' grounding is sufficient, a framework-based approach is highly efficient. For flagship products or domains requiring the highest possible precision and pedagogical depth, the investment in a custom, domain-aware pipeline is justified.

\subsection{Linguistic and Curricular Nuances in a Malaysian Context}
The study confirmed GPT-4o's proficiency in handling specific Bahasa Melayu mathematical terminology like integer, pecahan (fraction), perpuluhan (decimal), and tertib menurun (descending order) as used in the source documents.\cite{nota} The RAG methods were particularly effective at adopting the precise phrasing and style of the teacher-prepared notes. This, however, reveals a deeper nuance in the concept of ``curriculum alignment.'' The automated metrics used in this study excel at verifying topical and factual alignment. For example, the STS score confirms a question is about integers, and the RAG-QA check confirms it is answerable based on the curriculum's scope. Yet, the RPT specifies not just topics but also cognitive skills. Learning standard 1.2.6, ``Menyelesaikan masalah yang melibatkan integer'' (Solving problems involving integers), requires a higher cognitive level than standard 1.1.2, ``Mengenal dan memerihalkan integer'' (Recognizing and describing integers).\cite{nota} A simple definitional question might receive a high STS score for the topic of integers but completely fail to assess the problem-solving skill. This limitation of automated evaluation highlights that while these metrics serve as a powerful and scalable first-pass filter, they cannot fully capture the pedagogical or cognitive intent of a question. A complete validation framework must incorporate a final layer of human expert review to assess alignment with higher-order thinking skills.

\subsection{Resource Implications and Practicality}
The four methods have different resource profiles. Methods 1 and 2 are the least expensive in terms of API token consumption, as they involve a single call to the LLM. The RAG methods (3 and 4) are more costly, as they require additional API calls for embedding the source document chunks and the query. Method 4, with its custom processing, also incurs a higher computational cost during the chunking and retrieval stages. From a development perspective, the complexity increases from Method 2 (simplest) to Method 1, then Method 3, with Method 4 being the most intensive. This cost-benefit analysis suggests that for many practical applications, the significant quality jump from non-RAG to framework-based RAG (Method 3) represents the most favorable balance of cost, complexity, and performance.

\subsection{Evaluating the Evaluators}
A self-critical analysis of the evaluation framework is warranted. Its primary strengths are automation, scalability, and objectivity, allowing for the rapid assessment of hundreds of questions without human intervention. The introduction of the RAG-QA check as a functional validity test is a novel contribution that provides a stricter measure of alignment than similarity alone. However, as discussed, its primary limitation is its inability to reliably assess the cognitive level of a question. It can confirm what a question is about, but not how it asks it. This underscores the current frontier in AQG evaluation: moving from semantic correctness to pedagogical appropriateness.

\section{Conclusion and Future Work}

\subsection{Summary of Findings}
This study systematically investigated the use of Generative AI for creating curriculum-aligned multiple-choice questions for Form 1 Mathematics in Bahasa Melayu. The findings conclusively show that grounding the generation process using Retrieval-Augmented Generation (RAG) is not merely beneficial but essential for producing questions that are factually accurate and aligned with the specific learning standards of the Malaysian curriculum. RAG-based methods significantly outperformed non-grounded prompting techniques across all automated metrics. Furthermore, while a manual, domain-aware RAG pipeline offered the highest level of precision, a standard framework-based implementation provided a highly effective and practical solution, representing an optimal balance of performance and development effort for most use cases.

\subsection{Pathways to Scalability and Generalization}
The modular design of the proposed pipelines, particularly the RAG-based approaches, lends itself to scalability. The core methodology grounding an LLM in official curriculum documents is domain-agnostic. Future work will focus on applying this validated pipeline to other subjects within the KSSM framework, such as Sains (Science) and Sejarah (History), as well as other grade levels (e.g., Form 2, Form 3). The approach is also generalizable to other countries and low-resource language contexts that have structured curriculum documents but lack large-scale digital training data.

\subsection{The Essential Role of Human-in-the-Loop Validation}
This research underscores that AI should be a tool to augment, not replace, professional educators. While automated generation and evaluation can create a high-quality first draft of assessment materials, the final arbiter of pedagogical value remains the human teacher. Future work will involve a crucial human-in-the-loop validation study. The questions generated by the most effective pipeline (Method 4) will be presented to a panel of experienced Malaysian mathematics teachers for evaluation against a formal pedagogical rubric, potentially based on a framework like Bloom's Taxonomy, to assess cognitive complexity, clarity, and overall classroom suitability.\cite{ref5}

\subsection{Towards Personalization and Adaptive Learning}
The ultimate goal of creating a large, high-quality question bank is to power the next generation of educational technologies. The repository of generated questions can serve as the foundation for an adaptive learning system. Future research could explore extending the generation pipeline to create questions tailored to an individual student's learning profile, such as their demonstrated knowledge gaps or their preferred learning style (e.g., visual, auditory, reading/writing, kinesthetic - VARK).\cite{ref6} By dynamically generating questions of appropriate difficulty and style, such a system could create truly personalized learning pathways, realizing the full potential of AI to transform education in Malaysia and beyond.

\end{document}